\documentclass[journal,10pt]{IEEEtran}
\usepackage{multirow}
\usepackage{colortbl}
\usepackage{stfloats}
\usepackage[utf8]{inputenc} 
\usepackage[T1]{fontenc}    
\usepackage{url}            
\usepackage{booktabs}       
\usepackage{amsfonts}       
\usepackage{nicefrac}       
\usepackage{microtype}      

\usepackage[fleqn]{amsmath}
\usepackage{amsfonts}
\usepackage{amssymb}
\usepackage{bbm}  

\usepackage{graphicx}
\usepackage{amsthm}
\usepackage[noadjust]{cite}

\newtheorem{theorem}{Theorem}

\newtheorem{definition}[theorem]{Definition}

\newtheorem{lemma}[theorem]{Lemma}

\usepackage{array}

\usepackage[caption=false,font=footnotesize]{subfig}

\usepackage[bookmarks=false]{hyperref}

\usepackage{float}
\newfloat{footnote}{hb}

\DeclareMathOperator*{\argmin}{arg\,min}

\newcommand{\st}{\;\text{ s.t. }}

\newcommand{\SNR }{\text{SNR}}

\newcommand{\Null}{\textup{Null}}

\def \zero {{\mathbf 0 }}

\def \' {^{*}}

\def \A {{\boldsymbol A}}

\def \cOne    {{\mathcal{C}_0 }}
\def \cTwo    {{\mathcal{C}_1 }}

\def \E {{\mathbb E}}  
\def \e {{\boldsymbol e}}

\def \I {{\mathcal I}}

\def \P {{\mathbb P}}  

\def \R {\mathbb{R}}

\def \S {\mathcal{S}}

\def \sparse {s}

\def \v  {{\boldsymbol v}}

\def\x  {{\boldsymbol x}}
\def \optx { \boldsymbol{x^{\star}}}

\def\xB  {{\boldsymbol {x^B}}}

\def\xci  {{ \x_{\ci} }}

\def \y {{\boldsymbol y}}

\def \z {{\boldsymbol z}}

\def \zero {{\mathbf 0 }}

\def \loneminuo {\mathrm{ \mathbf{P_1^{{\lambda}}}}}

\def \loneminNoNoise {\mathrm{ \mathbf{P_1}}}

\def \ci  {j}

\def \xci                {{ \x_j }}
\def \yIci               {{ \y \text{\scriptsize $[\I_\ci]$} }}
\def \AIci              {{ \A{ [\I_{\ci}]}  }}

\def \AEi              {{ \A{ [\I]}  }}

\def \yEi               {{ \y{ \text{\scriptsize $[\I]$ } } }}

\def\zEi                 {{ \z \text{\scriptsize $[\I]$} }}

\def \yIcOne               {{ \y \text{\scriptsize $[\I_1]$} }}
\def \AIcOne              {{ \A{ [\I_{1}]}  }}

\def \yIcTwo               {{ \y \text{\scriptsize $[\I_2]$} }}
\def \AIcTwo              {{ \A{ [\I_{2}]}  }}

\def \yIcK               {{ \y \text{\scriptsize $[\I_K]$} }}
\def \AIcK              {{ \A{ [\I_{K}]}  }}

\title{Reducing Sampling Ratios and Increasing Number of Estimates Improve Bagging in Sparse Regression} 
\author{Luoluo~Liu$^J$,$\ $
          Sang~(Peter)~Chin$^{B, J} $, $\ $ 
          and Trac~D.~Tran$^J$\\ 
         lliu69@jhu.edu $~$ spchin@cs.bu.edu $~$ trac@jhu.edu\\
        $^J$ Department of Electrical Engineering, Johns Hopkins University, Baltimore, MD, 21210\\
        $^B$ Department of Computer Science $\&$ Hariri Institute of Computing, Boston University, Boston, MA, 02215
}
\begin{document}
%
\hypersetup{draft}
\pdfoutput=1
\maketitle
\addtolength{\belowcaptionskip}{-5mm}

\begin{abstract}
%
Bagging, a powerful ensemble method from machine learning, has shown the ability to improve the performance of unstable predictors in difficult practical settings. Although Bagging is most well-known for its application in classification problems, here we demonstrate that employing Bagging in sparse regression improves performance compared to the baseline method ($\ell_1$ minimization). Although the original Bagging method uses a bootstrap sampling ratio of $1$, such that the sizes of the bootstrap samples $L$ are the same as the total number of data points $m$, we generalize the bootstrap sampling ratio 
to explore the optimal sampling ratios for various cases.

The performance limits associated with different choices of bootstrap sampling ratio $L/m$ and number of estimates $K$ are analyzed theoretically.  Simulation results show that a lower $L/m$ ratio ($0.6 - 0.9$) leads to better performance than the conventional choice ($L/m = 1$), especially in challenging cases with low levels of measurements.  With the reduced sampling rate, SNR improves over the original Bagging method by up to $24\%$ and over the base algorithm $\ell_1$ minimization by up to $367 \%$. With a properly chosen sampling ratio, a reasonably small number of estimates ($K = 30$) gives a satisfying result, although increasing $K$ is discovered to always improve or at least maintain performance.\\

\end{abstract}

\vspace{-0.15cm}
\section{Introduction }
\label{sec:intro}
Compressed Sensing (CS) and Sparse Regression studies solving the linear inverse problem in the form of least squares with an additional sparsity-promoting penalty term. Formally speaking, the measurements vector $\y \in \R^m $ is generated by $\y = \A \x + \z$, where $\A \in \R^{m \times n}$ is the sensing matrix, $\x \in \R^n$ is a vector of sparse coefficients with very few non-zero entries, and $\z$ is a noise vector with bounded energy. The problem of interest is finding the sparse vector $\x$ given $\A$ as well as  $\y$. Among various choices of sparse regularizers, the $\ell_1$ norm is the most commonly used. The noiseless case is referred to as \emph{Basis Pursuit} (BP) whereas the noisy version is known as \emph{basis pursuit denoising}  \cite{bpdn}, or \emph{least absolute shrinkage and selection operator} (Lasso)~\cite{lasso}:
\begin{equation}
\label{eq:smv_ineq}
\loneminuo: \min  \lambda \| \x \|_1  + 0.5 \| \y - \A \x \|^2_2.
\end{equation}

The performance of $\ell_1$ minimization in recovering the true sparse solution has been thoroughly investigated in the CS literature~\cite{cs,CandesRobustUP,donohoCS,CSincoherence07}. CS theory reveals that if the sensing matrix $\A$ has good properties, then BP recovers the ground truth and the LASSO solution is close enough to the true solution with high probability~\cite{cs}.

Classical sparse regression recovery based on $\ell_1$ minimization solves the problem with all available measurements. In practice, it is often the case that not all measurements are available or required for recovery. Some measurements might be severely corrupted/missing or adversarial samples that break down the algorithm. These issues could lead to the failure of the sparse regression algorithm.

The Bagging procedure~\cite{bagging} proposed by 
Breiman is an efficient parallel ensemble method that improves the performance of unstable predictors. In Bagging, we first generate a bootstrap sample by randomly drawing $m$ samples uniformly with replacement from all $m$ data points.  We repeat the process $K$ times and generate $K$ bootstrap samples.
Then one bootstrapped estimator is computed for each bootstrap sample, and the final Bagged estimator is the average of all $K$ bootstrapped estimators.

Applying Bagging to find a sparse vector with a specific symmetric pattern was shown empirically to reduce estimation error when the sparsity level $\sparse$ is high~\cite{bagging} in a forward subset selection problem. This experiment shows the possibility of using Bagging to improve other sparse regression methods on general sparse signals. Although the well-known conventional Bagging method uses the bootstrap ratio $100\%$, some follow-up works have shown empirically that lower ratios improve Bagging in some classic classifiers: Nearest Neighbour Classifier~\cite{bagNN}, CART Trees~\cite{reducedSR}, Linear SVM, LDA, and Logistic Linear Classifier~\cite{reducedSRb}. Based on this success, we hypothesize that reducing the bootstrap ratio will also improve performance of Bagging in sparse regression. Therefore, we set up the framework with a generic bootstrap ratio and study its behavior with various bootstrap ratios.

In this paper, {\bf we use the notation $L$ as the sizes of bootstrap samples, $m$ as the number of all measurements, and $K$ as the number of estimates}.
{\it (i)} We demonstrate the generalized Bagging framework with bootstrap ratio $L/m$ and number of estimates $K$ as parameters.
{\it (ii)}  We explore the theoretical properties associated with finite $L/m$ and $K$.
{\it (iii)} We present simulation results with various parameters $L/m$ and $K$ and compare the performances of $\ell_1$ minimization, conventional Bagging, and Bolasso~\cite{bolasso}, another modern technique that incorporates Bagging into sparse recovery.
An important discovery is that in challenging cases with small $m$, Bagging with a ratio $L/m$ that is smaller than the conventional ratio $1$ can lead to better performance.

\section{Proposed Method: Bagging in Sparse Regression} 
\label{sec:jobs}

Our proposed method is sparse recovery using a generalized Bagging procedure. It is accomplished in three steps. First, {\bf we generate $K$ bootstrap samples, each of size $L$, randomly sampled uniformly and independently with replacement from the original $m$ data points. } This results
in $K$ measurements and sensing matrices pairs: $\{ \yIcOne , \AIcOne \} , \{\yIcTwo, \AIcTwo \} ...., \{ \yIcK, \AIcK \} $. We use the notation $(\cdot)[\I]$ on matrices or vectors to denote retaining only the rows supported on $\I$ and throwing away all other rows in the complement $\I^c$. 
Second, we solve the sparse recovery problem independently on each of those pairs; mathematically, for all $j = 1, 2,.., K$, we find
\begin{equation}
\label{eq:bmmv}
\x^{\boldsymbol{B}}_j =     \argmin_{\x \in \R^n }   \lambda_{(L,K)} \| \x \|_{1} + 0.5 \| \yIci  - \AIci  \x  \|^2_2, 
\end{equation}
where the parameter $\lambda_{(L,K)}$ is the balancing parameter of the least squares fit and the sparsity penalty for $(L,K)$ as the parameter choice for Bagging. The proposed approach (\ref{eq:bmmv}) is 
a Lasso problem, and numerous optimization methods can be used to solve it, such as~\cite{admm,spgl1Berg,spaRSA,liu2009}.

Finally, the Bagging solution is obtained by averaging all $K$ estimators from solving
(\ref{eq:bmmv}):
\begin{equation}
\mbox{ Bagging:} \quad \xB = \frac{1}{K}\sum_{\ci = 1}^K \x^{\boldsymbol{B}}_\ci.
\end{equation}

Compared to the $\ell_1$ minimization solution obtained from the usage of all the measurements,
the bagged solution $\xB$ is obtained by {\bf resampling without increasing the number of original measurements}. We will show that in some cases, the bagged solution outperforms the base $\ell_1$ minimization solution.

\section{Preliminaries}
\label{sec:prelim}

We summarize the theoretical results of CS theory which we need to analyze our algorithm mathematically.  We introduce the Null Space Property (NSP), as well as the Restricted Isometry Property (RIP). We also provide the tail bound of the sum of i.i.d. bounded random variables, which is needed to prove our theorems.

\subsection{Null Space Property (NSP)} 
The NSP~\cite{csNSP} for standard sparse recovery characterizes the necessary and sufficient conditions for successful sparse recovery using $\ell_1$ minimization.
\begin{theorem}[NSP]
\label{nsp_smv}
Every $\sparse-$sparse signal $\x \in \R^{n }$ is a unique solution to
$\loneminNoNoise : \ \min \| \x \|_{1} \st \y = \A \x $
if and only if $\A$ satisfies NSP of order $\sparse$.
Namely, if for all $\v \in \Null{(\A)} \backslash \{ \zero \}$, such that for any set $\S$ of cardinality less than or equals to the sparsity level $\sparse$
$ : \S \subset \{1,2,.., n \}, \text{card}(\S) \leq \sparse $, the following is satisfied:
\begin{equation*}
\|  \v \text{\footnotesize{$[\S]$ }} \|_{1} < \| \v \text{\footnotesize{$[\S^c]$ }} \|_{1},
\end{equation*}
where  $\v \text{\footnotesize{$[\S]$ }}$ only has the vector values
on an index set $\S$ and zero elsewhere.
\end{theorem}

\subsection{Restricted Isometry Property (RIP)} 
Although NSP directly characterizes the ability of success for sparse recovery, checking the NSP condition is computationally intractable. It is also not suitable to use NSP for quantifying performance in noisy conditions since it is a binary (True or False) metric instead of a continuous range. The Restricted isometry property (RIP)~\cite{cs} is introduced to overcome these difficulties. 
\begin{definition}[RIP]
\label{def:rip}
A matrix $\A$ with $\ell_2$-normalized columns satisfies RIP of order $\sparse$ if there exists a constant $\delta_{\sparse}(\A) \in [0 , 1) $ such that for every $\sparse-$sparse $\v \in \R^n$, the following is satisfied:
\vspace{-0.1in}
\begin{equation}
\label{eq:def_rip}
(1 - \delta_{\sparse} (\A) ) \|  \v \|_2^2 \leq  \| \A \v \|_2^2  \leq ( 1 + \delta_{\sparse}(\A) ) \| \v \|_2^2.
\end{equation}
\end{definition}

\subsection{Noisy Recovery bounds based on RIP constants}
It is known that satisfying the RIP conditions implies that the NSP conditions are also satisfied for 
sparse recovery~\cite{cs}. More specifically, if the RIP constant of order $2\sparse$ is strictly less than $\sqrt{2} - 1 $, then it implies that NSP is satisfied of the order $\sparse$.  
We recall Theorem 1.2 in~\cite{cs}, where the noisy recovery performance for $\ell_1$ minimization is bounded based on the RIP constant. This error bound is associated with the $\sparse-$sparse approximation error and the noise level.


\begin{theorem}[Noisy recovery for $\ell_1$ minimization~\cite{cs}] 
\label{th:noisy_recon_l1}
Let $ \y = \A \optx + \z$, $ \| \z \|_2 \leq \epsilon$, $\x_0$ is $\sparse-$sparse that minimizes $\| \x - \optx \|$ over all $\sparse-$sparse signals. If $\delta_{2 \sparse } (\A) \leq \delta < \sqrt{2} - 1$, $\x^{\boldsymbol{\ell_1}} $ be the solution of $\ell_{1} $ minimization
, then it obeys
\begin{equation*}
\| \x^{\boldsymbol{\ell_1}} - \optx \|_2 \leq \cOne(\delta ) \sparse^{-1/2} \| \x_0 - \optx\|_{1} + {\cTwo(\delta)} \epsilon ,
\end{equation*}
where $ \cOne(\cdot), \cTwo(\cdot)$ are some constants, which are determined by RIP constant $\delta_{2 \sparse}$. The form of these two constants terms are $\cOne(\delta) = \frac{ 2( 1 - (1 - \sqrt{2} )\delta)} {1- (1+ \sqrt{2})\delta }$ and $\cTwo (\delta) =  \frac{ 4 \sqrt{1 + \delta }}{ 1 - ( 1 + \sqrt{2} )  \delta }$.
\end{theorem}

\subsection{Tail bound of the sum of i.i.d. bounded Random variables}
This exponential bound is similar in structure to Hoeffidings' inequality. Proving this bound requires working with the moment generating function of a random variable.
\begin{lemma}[]
\label{lemma:sum_of_rvs}
Let $Y_1, Y_2,..., Y_n$ be i.i.d. observations of bounded random variable $Y$: $a \leq Y \leq b$ and the expectation $\E Y$ exists, for any $\xi > 0$, then
\begin{equation}
\label{eq:sum_bd_tail}
\P \{ \sum_{i = 1}^n Y_i \geq n \xi \} \leq \exp\{ - \frac{2 n ( \xi- \E Y) ^2}{(b - a)^2} \} .
\end{equation}
\end{lemma}

\section{Theoretical Results for Bagging associated with sampling ratio $L/m$ and the number of estimates $K$}
\subsection{Noisy Recovery for Employing Bagging in Sparse Regression}
We derive the performance bound for employing Bagging in sparse regression, 
in which the final estimate is the average over multiple estimates solved individually from bootstrap samples. We give the theoretical results for the case that true signal $\optx$ is exactly $\sparse-$sparse and the general case with no assumption of the sparsity level of the ground truth signal. Note that, the theorems are based on deterministic sensing matrix, measurements, and noise: $\A, \y, \z$, in which all vector norms are equivalent.

\begin{theorem}[Bagging: Error bound for $\|\optx \|_0 = \sparse $ ]
\label{th:noisy_bagging_exact_sparse}
Let $ \y = \A \optx + \z$, $ \| \z \|_2 < \infty $, 
If under the assumption that, for $\{ \I_\ci\}$s that generates a set of sensing matrices $\A{[\I_1]}, \A{[\I_2]},..., \A{[\I_K]}$,
there exists a constant that is relates to $L$ and $K$: $\delta_{(L,K)}$ such that for all $\ci \in \{1,2,...,K\}$, $\delta_{2 \sparse }(\A{[\I_\ci]}) \leq \delta_{(L,K)} < \sqrt{2} -1$. Let
$\xB $ be the solution of Bagging, then
for any $\tau > 0$, $\xB$ satisfies
\begin{equation*}
\label{eq:sum_upbd1}
\begin{split}
  \P    & \{   \| \xB- \optx \|_2  \leq
 \cTwo(\delta_{(L, K)})  (  \sqrt{ \frac{L}{m}}  \| \z \|_2 + \tau)  \}\\
   & \geq 1 - \exp \frac{- 2 K  \tau^4   }{ L^2 \| \z\|^4_\infty  } .
\end{split}
\end{equation*}
\end{theorem}


We also study the behavior of Bagging for a general signal $\optx, \| \optx \|_0 \geq \sparse $, in which the performance involves the $\sparse-$sparse approximation error. We use the vector $\e$ to denote this error, and $\e = \optx - \x_0$, where $\x_0$ is the best $\sparse$-sparse approximation of the ground truth signal over all $\sparse-$sparse signals. 

\begin{theorem}[Bagging: Error bound for general signal recovery]
\label{th:noisy_bagging_dm}
Let $ \y = \A \optx + \z$, $ \| \z \|_2 < \infty $,
If under the assumption that, for $\{ \I_\ci\}$s that generates a set of sensing matrices $\A{[\I_1]}, \A{[\I_2]},..., \A{[\I_K]}$, there exists $\delta_{(L,K)}$ such that for all $\ci \in \{1,2,...,K\}$, $\delta_{2 \sparse }(\A{[\I_\ci]}) \leq \delta_{(L,K)} < \sqrt{2} -1$. Let $\xB $ be the solution of Bagging, then
for any $\tau > 0$, $\xB$ satisfies
\begin{equation*}
\begin{split}
  \P & \{  \| \xB- \optx\|_2 \leq
\cOne(\delta_{L,K}) \sparse^{-1/2} \| \e \|_{1} + \\
& {\cTwo(\delta_{(L,K)})} (\sqrt{ \frac{L}{m}} \| \z \|_2 + \tau )  \}  \geq 1 - \exp  \frac{ - 2 K \cTwo^4(\delta_{(L,K)}) \tau^4}{ (b'    )^2} ,
\end{split}
\end{equation*}
where $b' = (\cOne(\delta_{(L,K)}) \sparse^{-1/2} \| \e\|_{1} + {\cTwo(\delta_{(L,K)})} \sqrt{L}\| \z \|_\infty)^{2} $.
\end{theorem}


Theorem~\ref{th:noisy_bagging_dm} gives the performance bound for Bagging in sparse signal recovery without the $\sparse-$sparse assumption, and it reduces to Theorem~\ref{th:noisy_bagging_exact_sparse} when the $\sparse-$sparse approximation error is zero $\|\e\|_1=0$.

We give the proof sketch that demonstrates the key idea to prove both Theorem~\ref{th:noisy_bagging_exact_sparse} and Theorem~\ref{th:noisy_bagging_dm}. The main tools are 
Theorem~\ref{th:noisy_recon_l1} and Lemma~\ref{lemma:sum_of_rvs}. Some special treatments are required to deal with terms while proving Theorem~\ref{th:noisy_bagging_dm}.
For more technical details, full proofs can be found in~\cite{jobs}.\\

\noindent \textbf{Proof Sketch:}
Similar to the sufficient condition in Theorem~\ref{th:noisy_recon_l1},  
the sufficient condition to analyze Bagging is that all matrices resulting from Bagging have well-behaved RIP constants of order $2s$ bounded by a universal constant $\delta$.

Let $\I$ denote a generic multi-set containing $L$ elements and each element in $\I$ is independent  and identically distributed, obeying a discrete uniform distribution from sample space $\{1,2,.., m\}$. The squared error function $f(\x \text{\scriptsize $(\I)$} )  = \| \x \text{\scriptsize $(\I)$} - \optx \|^2_2$, where $\x \text{\scriptsize $(\I)$}$ is the solution from $\ell_1$ minimization on $\I$: $\x \text{\scriptsize $(\I)$}= \argmin \| \x\|_1 \st \| \yEi - \AEi \|_2 \leq \epsilon_\I  $.  The squared errors from $K$ bootstrapped estimators $f(\xci) 
= \| \x^{\boldsymbol{B}}_\ci - \optx \|^2_2, j = 1, 2, ..., K$ are realizations generated i.i.d. from the distribution of $f(\x \text{\scriptsize $(\I)$} )$.

We proceed with the proof using Lemma~\ref{lemma:sum_of_rvs}. 
We choose the upper bound of the error to be a function of the expected value of noise power.  We pick the bound $\xi$ relating to the
the root of the expectation of squared error $ \sqrt{\E \|\zEi \|^2_2} = \sqrt{ \frac{L}{m}}  \| \z \|_2$. Then we need to compute the upper bound $b$ and the lower bound $a$ for the random variable $f( \x \text{\scriptsize $(\I)$}) $. Since it is non-negative, we choose $a = 0$. The upper bound $b$ is obtained from Theorem~\ref{th:noisy_recon_l1} and then the maximum value $\|\z\|_\infty$ is employed to further upper bound the noise level $\| \z  \text{\scriptsize{$[\I_j]$}} \|_2$. Through this process, we obtain the inequality: $ \P \{ \sum_{\ci} \| \x^{\boldsymbol{B}}_\ci - \optx \|^2_2 - K \xi \leq 0 \} \geq g(\E(f({\x}) , b ,a ) $, for some function $g$.

The Bagging solution is the average of all bootstrapped estimators. The key inequality to establish is as follows:
\begin{equation*}
\begin{split}
\P  & \{ \| \xB - \optx \|_2^2 - \xi  \leq 0 \} \\
= & \P \{ K \| \xB - \optx \|_2^2 - {\textstyle\sum}_{\ci} f({\xci })  + {\textstyle\sum}_{\ci} f({\xci }) - K \xi \leq 0 \} \\
 \geq & \P \{ K \| \xB - \optx \|_2^2  - {\textstyle\sum}_{\ci} f({\xci })   \leq 0 , {\textstyle\sum}_{\ci} f({\xci })  - K \xi \leq 0 \}\\
  =  & \P \{ K \| \xB - \optx \|_2^2  -  {\textstyle\sum}_{\ci} f({\xci })  \leq 0 \} \P \{ {\textstyle\sum}_{\ci} f({\xci }) - K \xi \leq 0 \}\\
  = & \P \{ {\textstyle\sum}_{\ci} \| \x^{\boldsymbol{B}}_\ci - \optx \|^2_2 - K \xi \leq 0 \}.
\end{split}
\end{equation*}
The first term is independent of the second term and it is true with probability $1$ by Jensens' inequality. Then we successfully establish the relationship of error bound of the Bagging solution to the sum of squared errors of bootstrapped estimates. To obtain the bound for the second term, we follow the method described in the previous paragraph.

\subsection{Parameters Selection Guided by the Theoretical Analysis}

Besides analyzing error bounds for general signals whose sparsity levels might exceed $\sparse$, Theorem~\ref{th:noisy_bagging_dm} can be used in analyzing cases 
when $m$ is not large enough for the sparsity level $\sparse$. 
Theorem~\ref{th:noisy_bagging_exact_sparse} and~\ref{th:noisy_bagging_dm} also guide us to optimal choices of parameters: the bootstrap sampling ratio $L/m$ and the number of estimates $K$.

Both Theorem~\ref{th:noisy_bagging_exact_sparse} and Theorem~\ref{th:noisy_bagging_dm}
show that increasing the number of estimates $K$ improves the result, by increasing the lower bound of certainty of the same performance. The growth rate of the certainty bound is decreasing with $K$. We validate this in our numerical experiment: even though increasing $K$ improves the results, the performance tends to be flattened out for a large $K$. 

The sampling ratio $L/m$ affects the result through two factors. The first one is the the RIP constant, which in general decreases with increasing $L$ (proved in~\cite{simpleRIP} with Gaussian assumption on sensing matrix). Since $\cTwo(\delta)$ is a non-decreasing function of $\delta$ and a larger $L$ usually results in a smaller $\delta$, then a larger $L$ in general results in a smaller $\cTwo(\delta)$. On the other hand, the second factor is the multiplier of  the noise power term, which is $\sqrt{L/m}$, suggesting a smaller $L$.

Combining these two factors indicates that the best $L/m$ ratio is somewhere in between a small and a large number. In the experiment results, we demonstrate that when $m$ is small, varying the bootstrap sampling ratio $L/m$ from $0-1$ creates peaks with the largest value at $L/m < 1$. The first factor, which relates $L$ to the RIP constant, is dominating in the stable case (when $m$ is sufficiently large), so that larger $L$ leads to better performance.
%

\begin{figure*}[t]
\centering
\subfloat[$m = 50$]{\includegraphics[trim = {0 0 0 20}, clip, width=0.245\textwidth]{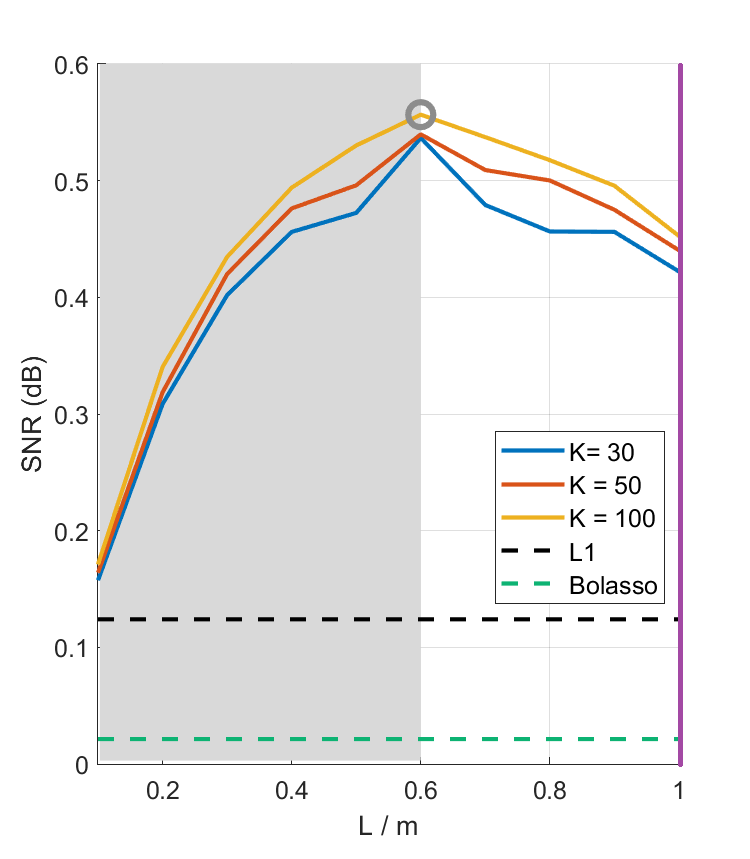}}
\subfloat[$m = 75$]{\label{fg:l12}\includegraphics[trim = {0 0 0 20}, clip, width=0.245\textwidth]{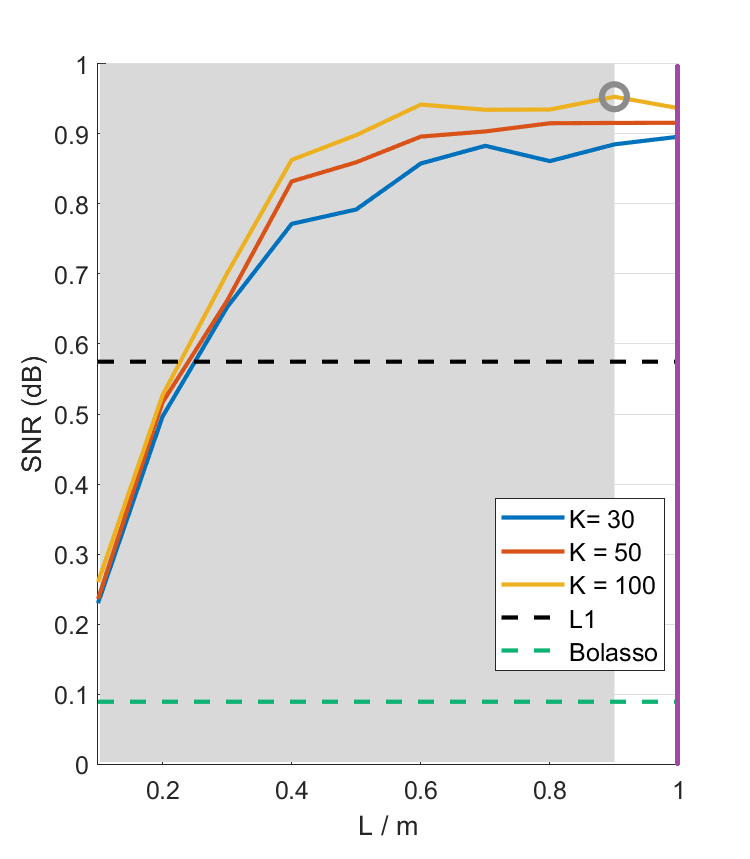}}
\subfloat[$m = 100$]{\label{fg:l13}\includegraphics[trim = {0 0 0 20}, clip, width=0.245\textwidth]{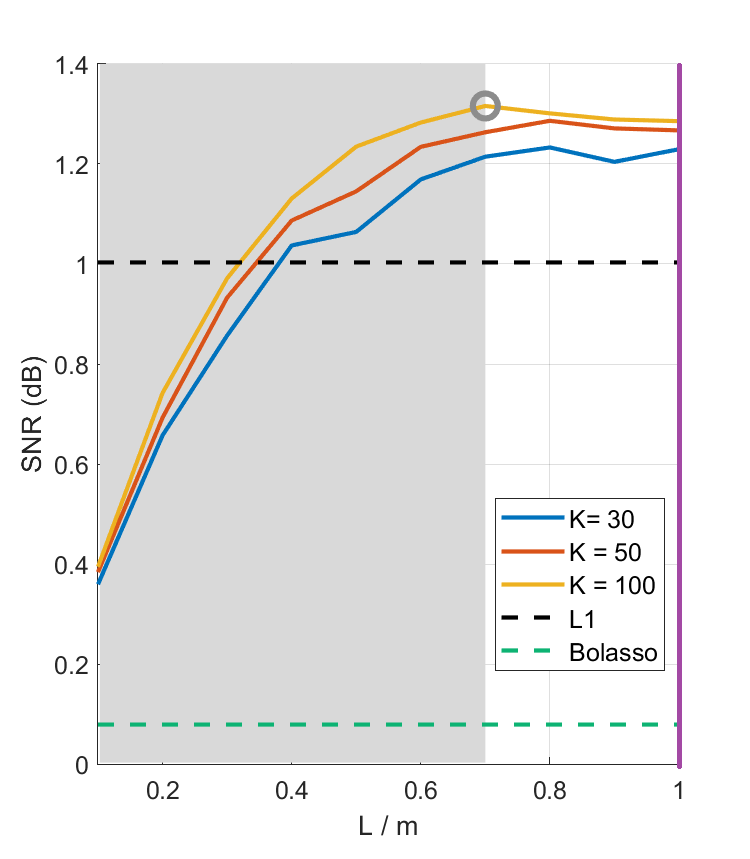}}
\subfloat[$m = 150$]{\label{fg:l14}\includegraphics[trim = {0 0 0 20}, clip, width=0.245\textwidth]{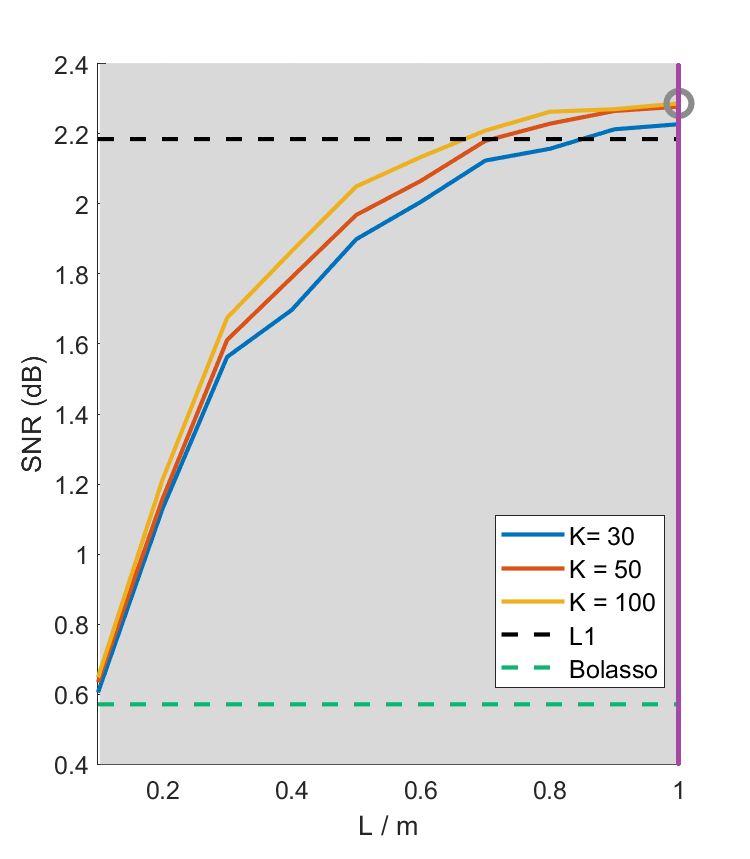}}

\caption{Performance curves for Bagging 
with various sampling ratios $L/m$ and number of estimates $K$, the best performance of Bolasso as well as $\ell_1$ minimization. The Purple lines highlighted conventional Bagging with $L/m =1$.
In all cases, $\SNR=0 $ dB and the number of measurements $m = 50 , 75, 100, 150$ from left to right.
The grey circle highlights the peak of Bagging
, and the grey area highlights the bootstrap ratio at the peak point. }
\vspace{-0.1cm}
\label{fg:exp1}
\end{figure*}

\begin{table*}[bp]
\centering
\caption{The performance of $\ell_1$ minimization and the best performance among all choices of $L$ and $K$  for Bagging, Bolasso methods with various total number of measurements $m$.
SNR $= 0  \mbox{{d}B} $. $\ $  All performances are measured by the averaged recovered SNR (dB) }
\begin{tabular}{|c||>{\columncolor[gray]{0.9}}c|>{\columncolor[gray]{0.9}}c|>{\columncolor[gray]{0.9}}c|c|c|c|c|>{\columncolor[gray]{0.92}}c|>{\columncolor[gray]{0.92}}c|c| }
  \hline
 & \multicolumn{3}{c|}{\cellcolor[gray]{0.9} Small $m$} & \multicolumn{4}{c|}{ Moderate $m$}  & \multicolumn{2}{c|}{\cellcolor[gray]{0.92} Large $m$} & Very large $m$ \\
The number of measurements $m$    &  50 &  75 &  100 &  125 &  150 &  175 &  200 &  500 & 1000&  2000\\
\hline
\hline
$\ell_1$ min.               & 0.12  & 0.57 & 1.00 & \underline{1.70} & \underline{2.19} & \underline{2.61} & \underline{2.97} & {\bf 6.53} & {\bf 9.46} & \underline{12.55} \\
Conventional Bagging (L/m=1)    & \underline{0.45} & \underline{0.94} & \underline{1.29} & {\bf 1.86} & {\bf 2.29} & {\bf 2.70} & {\bf 3.01} & \underline{6.22} & \underline{9.06} & 12.10 \\
Bagging                     & {\bf 0.56} & {\bf 0.95} & {\bf 1.32} & {\bf 1.86} & {\bf 2.29} & {\bf 2.70} & {\bf 3.01} & \underline{6.22} & \underline{9.06} & 12.10\\
Bolasso                     & 0.02 & 0.09 & 0.08 & 0.28 & 0.57 & 0.98 & 1.23 & 5.21 & 8.94 & {\bf 12.73}  \\
  \hline
\end{tabular}
\label{tb:pv_snr0}
\end{table*}

\section{Simulations }  
\label{sec:simulation}
In this section, we perform sparse recovery on simulated data to study the performance of our algorithm. In our experiment, all entries of $ \A \in \R^{m \times n}$ are i.i.d. samples from the standard normal distribution $\mathcal{N} (0, 1)$. 
The signal dimension $ n = 200 $ and various numbers of measurements from $50$ to $2000$ are explored. 
For the ground truth signals, their sparsity levels are all $\sparse=50$, and the non-zero entries are sampled from the standard Gaussian with their locations being generated uniformly at random.
For the noise processes $\z$, entries are sampled i.i.d. from $\mathcal{N} ( 0 , \sigma^2) $, with variance $\sigma^2 = 10^{-\SNR/10} \| \A \x \|_2^2$, where $\SNR$ represents the Signal to Noise Ratio.
We add white Gaussian noise to make the $\SNR = 0$ dB. All numerical realizations have finite values.
We use the ADMM~\cite{admm} implementation of Lasso to solve all sparse regression problems, in which the parameter $\lambda_{(L,K)}$ balances the least squares fit and the sparsity penalty for the case with $(L,K)$ as parameters.

We study how the bootstrap sampling ratio $L/m$ as well as the number of estimates $K$ affects the result. In our experiment, we take $K = 30, 50, 100$ and $L/m $ from $0.1$ to $1$. 
We report the Signal to Noise Ratio (SNR) as the error measure for recovery: $ \SNR (\x , \optx )= 10 \log_{10} \| \x - \optx \|_2^2 / \| \optx \|_2^2$ averaged over $20$ independent trials. For all algorithms, we evaluate $\lambda_{(L,K)}$ at different values from $.01$ to $200$ and then select optimal values that give the maximum averaged SNR over all trials.

\subsection{Performance of Bagging, Bolasso and $\ell_1$ minimization}
Bagging and Bolasso with the various parameters $K,L$ and $\ell_1$ minimization are studied.  The results are plotted in Figure~\ref{fg:exp1}.
 The colored curves show the cases of Bagging with various number of estimates $K$. The intersections of colored curves and the purple solid vertical lines at $L/m =1$ illustrates  conventional Bagging with a full bootstrap rate. The grey circle highlights the best performance and the grey area highlights the optimal bootstrap ratio $L/m$.
 The performance of $\ell_1$ minimization is depicted by the black dashed lines, while the best Bolasso performance is plotted using light green dashed lines. In those figures, for each condition with a choice of $L,K$, the information available to Bagging and Bolasso algorithms are identical, and $\ell_1$ minimization always has access to all $m$ measurements.

From Figure~\ref{fg:exp1}, we see that when $m$ is small, Bagging can outperform $\ell_1$ minimization. As $m$ decreases, the margin increases. 
The important observation is that 
when the number of measurements is low ($m$  is between $\sparse$ to $2\sparse$: $50 - 100$, $\sparse$ is the sparsity level),
by using a reduced bootstrap ratio $L/m$ ($60\% - 90 \%$), Bagging beats the conventional choice of the full ratio $1$ for all different choices of $K$. Also with a reduced ratio and a small $K$
our algorithm is already quite robust and outperforms $\ell_1$ minimization by a large margin.
When the number of measurements is moderate $m  = 3 \sparse = 150$, Bagging still beats the baseline; however,  the optimal parameters here are bootstrap ratio
$L/m =1$ and the number of estimates $K = 100$.
In this case, the reduced bootstrap ratio does not bring any performance improvement.
Increasing the level measurement makes the base algorithm more stable and the advantage of Bagging starts decaying.


We perform the same experiments with higher number of measurements $m$, and Table~\ref{tb:pv_snr0} illustrates the best performance for various schemes: $\ell_1$ minimization, the original Bagging scheme with a full bootstrap ratio, Bagging, and Bolasso 
with $\SNR = 0$ dB.
For Bagging, 
 the peak values are found among different choices of parameters $K$ and $L$ that we explored. We see that when the number of measurements $m$ is small ($50-100$), Bagging outperforms $\ell_1$ minimization. The reduced bootstrap rate also improves conventional Bagging: the improvement is significant: $24\%$ on SNR when $m = 50$.  When $m$ is moderate ($125-200$),
choosing reduced rates does not improve the performance compared to conventional Bagging. Bagging still outperforms
$\ell_1$ minimization
with smaller margins than the cases with small $m$. 
While $m$ is large ($ \geq 500$), Bagging starts losing its advantage over $\ell_1$ minimization. 
Bolasso 
only performs similarly to other algorithms in the easiest case for an extremely large $m$ ($=2000$) where it slightly outperforms all other algorithms. 
\section{Conclusion}
We extend the conventional Bagging scheme in sparse recovery with the bootstrap sampling ratio $L/m$ as adjustable parameters
and derive error bounds for the algorithm associated with $L/m$ and the number of estimates $K$. 
Bagging is particularly powerful when the number of measurements $m$ is small. Although this condition is notoriously difficult, both in terms of improving sparse recovery results and obtaining tight bounds of theoretical properties, 
Bagging outperforms $\ell_1$ minimization by a large margin (up to 367\%). Moreover, the reduced sampling rate shows a performance improvement measured by the recovered SNR, and it is over the conventional Bagging algorithm 
by up to $24\%$.

Our Bagging scheme achieves acceptable performance even with very small $L/m$ (around $0.6$) 
and relative small $K$ (around $30$ in our experimental study).
The error bounds for Bagging predict that a smaller sampling rate $L/m$ can  lead to performance improvement and
increasing $K$ improves the certainty of the bound. Both are validated in our numerical simulation.
For a sequential system, a reasonably large $K$ (around $30$) is enough to obtain an fairly good solution.
For a parallel system that allows a large amount of processes to be run at the same time, 
a large $K$ is preferred since it
 in general gives a better result.
\newpage
\section{Acknowledgement}
We would like to thank Dr. Dror Baron for insightful comments and suggestions, Dr. Cindy Rush for thoughtful feedbacks, and Nicholas Huang for efforts in helping polish,
all towards improving the overall quality of our paper.\\

\bibliographystyle{unsrt}
 \bibliography{ieee_bagging}

%
%
%
%
%
\end{document}